\documentclass{article}
\usepackage{ismir,amsmath,cite,url}
\usepackage{graphicx}
\usepackage{color}
\usepackage{booktabs}

\newcommand{\subf}[2]{%
  {\small\begin{tabular}[t]{@{}c@{}}
  #1\\#2
  \end{tabular}}%
}

\mathchardef\mhyphen="2D 
\usepackage{flushend} 

\apptocmd{\thebibliography}{%
  \setlength{\itemsep}{2pt}%
  \setlength{\parskip}{4pt}%
}{}{}

\title{Deep Watershed Detector for Music Object Recognition}

\multauthor
{Lukas Tuggener$^{1,3}$ \hspace{0.5cm} Ismail Elezi$^{1,2}$  } { \bfseries{J{\"u}rgen Schmidhuber$^3$ \hspace{0.5cm} Thilo Stadelmann$^1$}\\
$^1$ ZHAW Datalab, Zurich University of Applied Sciences, Winterthur, Switzerland\\
$^2$ Dept. of Environmental Sciences, Informatics and Statistics, Ca'Foscari University of Venice, Italy\\
$^3$ Faculty of Informatics, Universit\`{a} della Svizzera Italiana, Lugano, Switzerland\\
{\tt\small tugg@zhaw.ch, ismail.elezi@unive.it, juergen@idsia.ch, stdm@zhaw.ch}
}
\def\authorname{Lukas Tuggener, Ismail Elezi, J{\"u}rgen Schmidhuber, Thilo Stadelmann}

\begin{document}

\maketitle
\begin{abstract}
Optical Music Recognition (OMR) is an important and challenging area within music information retrieval, the accurate detection of music symbols in digital images is a core functionality of any OMR pipeline. In this paper, we introduce a novel object detection method, based on synthetic energy maps and the watershed transform, called Deep Watershed Detector (DWD). Our method is specifically tailored to deal with high resolution images that contain a large number of very small objects and is therefore able to process full pages of written music. We present state-of-the-art detection results of common music symbols and show DWD's ability to work with synthetic scores equally well as on handwritten music.
\end{abstract}
\section{Introduction and Problem Statement}\label{sec:introduction}
The goal of Optical Music Recognition (OMR) is to transform images of printed or handwritten music scores into machine readable form, thereby understand the semantic meaning of music notation \cite{omr}. It is an important and actively researched area within the music information retrieval community. 
The two main challenges of OMR are: first the accurate detection and classification of music objects in digital images; and second, the reconstruction of valid music in some digital format. This work is focusing solely on the first task.

Recent progress in computer vision \cite{Detectron2018} thanks to the adaptation of convolutional neural networks (CNNs) \cite{DBLP:journals/pr/FukushimaM82, DBLP:journals/neco/LeCunBDHHHJ89} provide a solid foundation for the assumption that OMR systems can be drastically improved by using CNNs as well. Initial results of applying deep learning \cite{DBLP:journals/nature/LeCunBH15, DBLP:journals/nn/Schmidhuber15} to heavily restricted settings such as staffline removal \cite{DBLP:journals/eswa/GallegoC17}, symbol classification \cite{DBLP:conf/icmla/PachaE17} or end-to-end OMR for monophonic scores \cite{DBLP:conf/ismir/Calvo-ZaragozaV17}, support such expectations. 

In this paper, we introduce a novel general object detection method called Deep Watershed Detector (DWD) motivated by the following two hypotheses: a) deep learning can be used to overcome the classical OMR approach of having hand-crafted pipelines of many preprocessing steps \cite{DBLP:journals/ijdar/RebeloCC10} by being able to operate in a fully data-driven fashion; b) deep learning can cope with larger, more complex inputs than simple glyphs, thereby learning to recognize musical symbols in their context. This will disambiguate meanings (e.g., between staccato and augmentation dots) and allow the system to directly detect a complex alphabet. 

DWD operates on full pages of music scores in one pass without any preprocessing besides interline normalization, detects handwritten and digitally rendered music symbols without any restriction on the alphabet of symbols to be detected. We further show that it learns meaningful representation of music notation and achieves state-of-the art detection rates on common symbols. 

The remaining structure of this paper is as follows: Sec. \ref{sec:related} puts our approach in context with existing methods; in Sec. \ref{sec:methods} we derive our original end-to-end model, and give a detailed explanation on how we use the deep watershed transform for the task of object recognition; Sec. \ref{sec:experiments} reports on experimental results of our system on the \emph{DeepScores} digitally rendered dataset in addition to the \emph{MUSCIMA++} handwritten dataset before we conclude in Sec. \ref{sec:conc} with a discussion and give pointers for future research. 

\begin{figure*}
\centering
\includegraphics[scale=0.3]{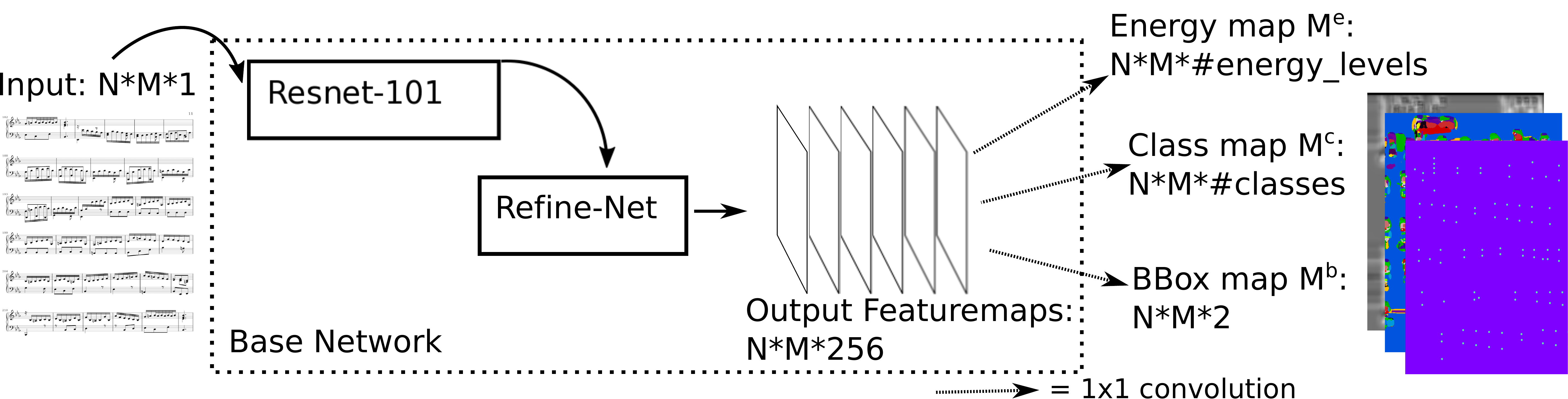}
\caption{Illustration of the DWD network and its sub-components together with input and outputs. The outputs have been cropped to improve visibility} 
\label{fig:network}
\end{figure*}

\section{Related Work}\label{sec:related}
The visual detection and recognition of objects is one of the most central problems in the field of computer vision. With the recent developments of CNNs, many competing CNN-based approaches have been proposed to solve the problem. R-CNNs \cite{DBLP:conf/cvpr/GirshickDDM14}, and in particular their successors \cite{DBLP:conf/nips/RenHGS15}, are generally considered to be state-of-the-art models in object recognition, and many developed recognition systems are R-CNN based. On the other hand, researchers have also proposed models which are tailored towards computational efficiency instead of detection accuracy. YOLO systems \cite{DBLP:conf/cvpr/RedmonF17} and Single-Shot Detectors \cite{DBLP:conf/eccv/LiuAESRFB16} while slightly compromising on accuracy, are significantly faster than R-CNN models, and can even achieve super real-time performance.

A common aspect of the above-mentioned methods is that they are specifically developed to work on cases where the images are relatively small, and where images contain a small number of relatively large objects \cite{DBLP:journals/ijcv/EveringhamGWWZ10, DBLP:conf/eccv/LinMBHPRDZ14}. On the contrary, musical sheets usually have high-resolution, and contain a very large number of very small objects, making the mentioned methods not suitable for the task.

The watershed transform is a well understood method that has been applied to segmentation for decades \cite{beucher1992watershed}. 
Bai and Urtasun \cite{DBLP:conf/cvpr/BaiU17} were first to propose combining the strengths of deep learning with the power of this classical method. They proposed to directly learn the energy for the watershed transform such that all dividing ridges are at the same height. As a consequence, the components can be extracted by a cut at a single energy level without leading to over-segmentation. The model has been shown to achieve state of the art performance on object segmentation.

For the most part, OMR detectors have been rule based systems working well only within a hard set of constraints \cite{DBLP:journals/ijdar/RebeloCC10}. Typically, they require domain knowledge, and work well only on simple typeset music scores with a known music font, and a relatively small number of classes \cite{DBLP:journals/ejasp/RossantB07}. When faced with low-quality images, complex or even handwritten scores \cite{DBLP:conf/icfhr/BaroRF16}, the performance of these models quickly degrades, to some degree because errors propagate from one step to another \cite{DBLP:conf/icmla/PachaE17}. Additionally, it isn't clear what to do when the classes change, and in many cases, this requires building the new model from scratch.

In response to the above mentioned issues some deep learning based, data driven approaches have been developed. Hajic and Pecina \cite{DBLP:journals/corr/abs-1708-01806} proposed an adaptation of Faster R-CNN with a custom region proposal mechanism based on the morphological skeleton to accurately detect noteheads, while Choi et al. \cite{DBLP:conf/icdar/ChoiCRZ17} were able to detect accidentals in dense piano scores
with high accuracy, given previously detected noteheads, that are being used as input-features to the network. A big limitation of both approaches is that the experiments have been done only on a tiny vocabulary of the musical symbols, and therefore their scalability remains an open question.

To our knowledge, the best results so far has been reported in the work of Pacha and Choi \cite{DBLP:conf/iwdas/Pacha2018} where they explored many models on the \emph{MUSCIMA++ }\cite{DBLP:conf/icdar/HajicP17b} dataset of handwritten music notation. They got the best results with a Faster R-CNN model, achieving an impressive score on the standard mAP metric. A serious limitation of that work is that the system wasn't designed in an end-to-end fashion and needs heavy pre- and post-processing. In particular, they cropped the images in a context-sensitive way, by cutting images first vertically and then horizontally, such that each image contains exactly one staff and has a width-to-height-ratio
of no more than $2:$1, with about $15\%$ horizontal overlap to adjacent slices. In practice, this means that all objects significantly exceeding the size of such a cropped region will neither  appear in the training nor testing data, as only annotations that have an intersection-over-area of $0.8$ or higher between the object and the cropped region are considered part of the ground truth. Furthermore, all the intermediate results must be combined to one concise final prediction, which is a non-trivial task.

\section{Deep Watershed Detection}\label{sec:methods}

\begin{figure}
\centering
\begin{tabular}{c}

\subf{\includegraphics[width=70mm]{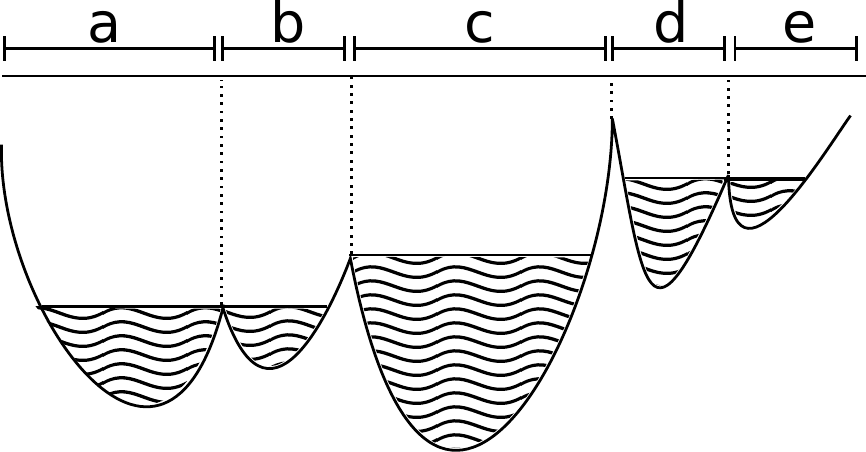}}
     {\textbf{a)} One-dimensional energy function of five \\ classes without any structural constraints.}
\\
\subf{\includegraphics[width=70mm]{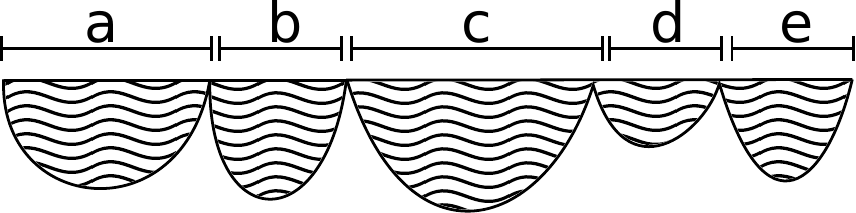}}
     {\textbf{b)} Energy function for the same five classes \\ with fixed boundary energy.}
\\

\subf{\includegraphics[width=70mm]{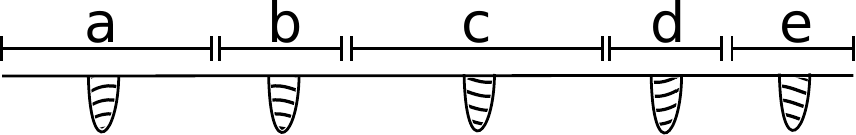}}
     {\textbf{c)} Energy function for the same five classes this time \\  with small energy markers at the class centers.}
\\
\end{tabular}
\caption{Illustration of the watershed transform applied to different one-dimensional functions.}
\label{fig:watershed_energy}
\end{figure}

In this section we present the Deep Watershed Detector (DWD) as a novel object detection system, built on the idea of the deep watershed transform \cite{DBLP:conf/cvpr/BaiU17}. The watershed transform \cite{beucher1992watershed} is a mathematically well understood method with a simple core idea that can be applied to any topological surface. The algorithm starts filling up the surface from all the local minima, with all the resulting basins corresponding to connected regions. When applied to image gradients, the basins correspond to homogeneous regions of said image (see Fig. \ref{fig:watershed_energy}a). One key drawback of the watershed transform is its tendency to over segment. This issue can be addressed by using the deep watershed transform. It combines the classical method with deep learning by training a deep neural network to create an energy surface based on an input image. This has the advantage that one can design the energy surface to have certain properties. When designed in a way that all segmentation boundaries have energy zero, the watershed transform is reduced to a simple cutoff at a fixed energy level (see Fig. \ref{fig:watershed_energy}b). An objectness energy of this fashion has been used by Bai and Urtasun for instance segmentation \cite{DBLP:conf/cvpr/BaiU17}. Since we want to do object detection, we further simplify the desired energy surface to having small conical energy peaks of radius $n$ pixels at the center of each object and be zero everywhere else (see Fig. \ref{fig:watershed_energy}c).

More formally, we define our energy surface (or: energy map) $M^{e}$ as follows:

\begin{equation}\label{energy_map}
M^{e}_{(i,j)} = max \begin{cases}
    \ \underset{c \in C}{\mathrm{argmax}}[E_{max}\cdot(1-\frac{\sqrt{(i-c_i)^2+(j-c_j)^2}}{r})]\\
    0 
    \end{cases}
\end{equation}

where $M^{e}_{(i,j)}$ is the value of $M^e$ at position $(i,j)$, $C$ is the set of all object centers and $c_i, c_j$ are the center coordinates of a given center $c$. $E_{max}$ corresponds to the maximum energy and $r$ is the radius of the center marking.

At first glance this definition might lead to the misinterpretation that object centers that are closer together than $r$ cannot be disambiguated using the watershed transform on $M^e$. This is not the case since we can cut the energy map at any given energy level between $1$ and $E_{max}$. However, using this method it is not possible to detect multiple bounding boxes that share the exact same center.

\subsection{Retrieving Object Centers}
After computing an estimate $\hat M^e$ of the energy map, we retrieve the coordinates of detected objects by the following steps:
\begin{enumerate}
\item Cut the energy map at a certain fixed energy level and then binarize the result.
\item Label the resulting connected components, using the two-pass algorithm \cite{Wu:2009:OTC:1529867.1529869}. Every component receives a label $l$ in $1...n$, for every component $o^l$ we define $O^l_{ind}$ as the set of all tuples $(i,j)$ for which the pixel with coordinates $j$ and $i$ is part of $o^l.$ 
\item The center $\hat c^l$ of any component $o^l$ is given by its center of gravity: 
\begin{equation}\label{center}
\hat c^l = o^l_{center} = \vert O^l_{ind} \vert^{-1}  \cdot \sum_{(i,j) \in O^l_{ind}}{(i,j)}
\end{equation}
\end{enumerate}
We use these component centers $\hat c$ as estimates for the object centers $c$.

\subsection{Object Class and Bounding Box}
In order to recover bounding boxes we do not only need the object centers, but also the object classes and bounding box dimensions. To achieve this we output two additional maps $M^c$ and $M^b$ as predictions of our network. $M^c$ is defined as:

\begin{equation}\label{Mc}
M^{c}_{(i,j)} = \begin{cases}
    \Lambda_{(i,j)},& \text{if } M^{e}_{(i,j)} > 0\\
    \Lambda_{background} ,  & \text{otherwise}
\end{cases}
\end{equation}

where $\Lambda_{backgroud}$ is the class label indicating background and $\Lambda_{(i,j)}$ is the class label associated with the center $c$ that is closest to $(i,j)$. We define our estimate for the class of component $o^l$ by a majority vote of all values $\hat M^{c}_{(i,j)}$ for all $(i,j) \in O^l_{ind}$, where $\hat M^c$ is the estimate of $M^c$. Finally, we define the bounding box map $M^{b}$ as follows:
\begin{equation}\label{bbox_map}
M^{b}_{(i,j)} = \begin{cases}
    (y^{l}, x^{l}),& \text{if } M^{e}_{(i,j)} > 0\\
    (0,0) ,  & \text{otherwise}
\end{cases}
\end{equation}
where $y^{l}$ and $x^{l}$ are the width and height of the bounding box for component $o^l$. Based on this we define our bounding box estimation as the average of all estimations for label $l$: 
\begin{equation}\label{bbox_estimation}
(\hat y^l, \hat x^l) = \vert O^l_{ind} \vert^{-1} \cdot\sum_{(i,j) \in O^l_{ind}}{\hat M^b_{(i,j)}}
\end{equation}

\begin{figure*}
\centering
\begin{tabular}{c}

\subf{\includegraphics[scale=0.35]{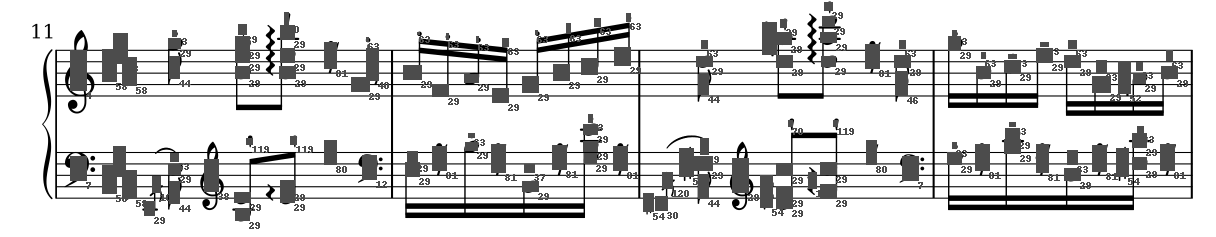}}
     {\textbf{a)} Example result from \emph{DeepScores} with detected bounding boxes as overlays. The tiny numbers are class labels from the \\ dataset introduced with the overlay. This system is roughly one forth of the size of a typical \emph{DeepScores} input we process at once.}
\\
\subf{\includegraphics[scale=0.35]{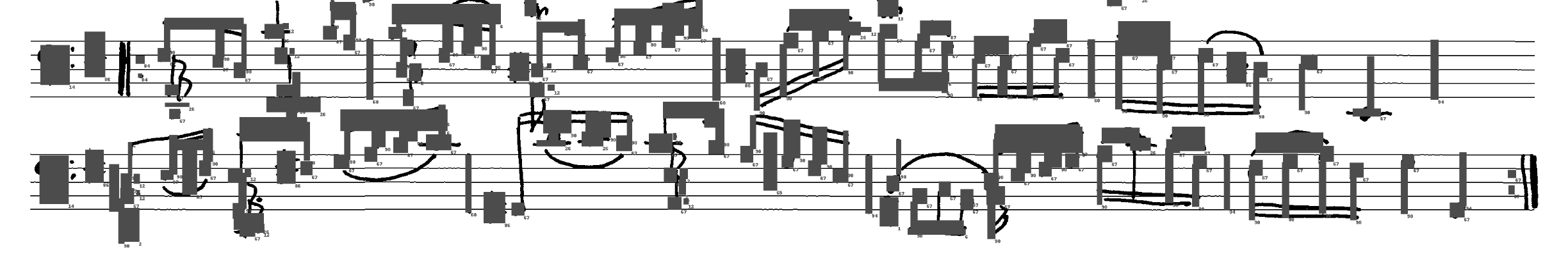}}
     {\textbf{b)} Example result from \emph{MUSCIMA++} with detected bounding boxes and class labels as overlays. This system is roughly \\ one half of the size of a typical processed \emph{MUSCIMA++} input. The images are random picks amongst inputs with many symbols.}
\\
\end{tabular}
\caption{Detection results for \emph{MUSCIMA++} and \emph{DeepScores} examples, drawn on crops from corresponding input images.}
\label{fig:ex_detection}
\end{figure*}

\subsection{Network Architecture and Losses}\label{subsec:architecture}
As mentioned above we use a deep neural network to predict the dense output maps $M^e$, $M^c$ and $M^b$ (see Fig. \ref{fig:network}). The base neural network for this prediction can be any fully convolutional network with the same input and output dimensions. We use a ResNet-101 \cite{DBLP:conf/cvpr/HeZRS16} (a special case of a Highway Net \cite{greff2015nips}) in conjunction with the elaborate RefineNet \cite{DBLP:conf/cvpr/LinMSR17} upsampling architecture. For the estimators defined above it is crucial to have the highest spacial prediction resolution possible. Our network has three output layers, all of which are an $1$ by $1$ convolution applied to the last feature map of the RefineNet.

\subsubsection{Energy prediction} We predict a quantized and one-hot encoded version of $M^e$, called $M^{e \mhyphen oh}$, by applying a 1 by 1 convolution of depth $E_{max}$ to the last feature map of the base network. The loss of the prediction $\hat M^{e \mhyphen oh}$, $loss^e$, is defined as the cross-entropy between $M^{e \mhyphen oh}$ and $\hat M^{e \mhyphen oh}$.

\subsubsection{Class prediction} We again use the corresponding one-hot encoded version $M^{c \mhyphen oh}$ and predict it using an $1$ by $1$ convolution, with the depth equal to the number of classes, on the last feature map of the base network. The cross-entropy $loss^c$ is calculated between between $M^{c \mhyphen oh}$ and $\hat M^{c \mhyphen oh}$. Since it is not the goal of this prediction to distinguish between foreground and background, all the loss stemming from locations with $M^e=0$ will get masked out.

\subsubsection{Bounding box prediction} $M^b$ is predicted in its initial form using an $1$ by $1$ convolution of depth $2$ on the last feature map of the base network. The bounding box loss $loss^b$ is the mean-squared difference between $M^b$ and $\hat M^b$. For $loss^b$, the components stemming from background locations will be masked out analogous to $loss^c$.

\subsubsection{Combined prediction}
We want to jointly train in all tasks, therefore we define a total loss $loss^{tot}$ as: \begin{equation}\label{loss_tot}
loss^{tot} = w_1 * \frac{loss^e}{v^e} + w_2 * \frac{loss^c}{v^c} + w_3 * \frac{loss^b}{v^b}
\end{equation}
where the $v^.$ are running means of the corresponding losses and the scalars $w_.$ are hyper-parameters of the DWD network. We purposefully use very short extraction heads of one convolutional layer; by doing so we force the base network to do all three tasks simultaneously. We expect this leads to the base network learning a meaningful representation of music notation, from which it can extract the solutions of the three above defined tasks. 

\section{Experiments and Results}\label{sec:experiments}

\subsection{Used Datasets}\label{subsec:datasets}

For our experiments we use two datasets: \emph{DeepScores} \cite{DeepScores} and \emph{MUSCIMA++} \cite{DBLP:conf/icdar/HajicP17b}.

\emph{DeepScores} is currently the largest publicly available dataset of musical sheets with ground truth for various machine learning tasks, consisting of high-quality pages of written music, rendered at $400$ dots per inch. The dataset has $300,000$ full pages as images, containing tens of millions of objects, separated in $123$ classes. We randomly split the set into training and testing, using $200k$ images for training and $50k$ images each for testing and validation. The dataset being so large allows efficient training of large convolutional neural networks, in addition to being suitable for transfer learning \cite{DBLP:conf/nips/YosinskiCBL14}.

\emph{MUSCIMA++} is a dataset of handwritten music notation for musical symbol detection. It contains $91,255$ symbols spread unto $140$ pages, consisting of both notation primitives and higher-level notation objects, such as key signatures or time signatures. It features 105 object classes. There are $23,352$ notes in the dataset, of which $21,356$ have a full notehead, $1,648$ have an empty notehead, and $348$ are grace notes. We randomly split the dataset into training, validation and testing, with the training set consisting of $110$ pages, while validation and testing each consist of $15$ pages.

\subsection{Network Training and Experimental Setup}
We pre-train our network in two stages in order to achieve reasonable results. First we train the ResNet on music symbol classification using the \emph{DeepScores} classification dataset \cite{DeepScores}. Then, we train the ResNet and RefineNet jointly on semantic segmentation data also available from \emph{DeepScores}. After this pre-training stage we are able to use the network on the tasks defined above in Sec. \ref{subsec:architecture}.

Since music notation is composed of hierarchically organized sub-symbols, there does not exist a canonical way to define a set of atomic symbols to be detected (e.g., individual numbers in time signatures vs. complete time signatures). We address this issue using a fully data driven approach and detecting the unaltered labels as they are provided by the two datasets.

We rescale every input image to the desired interline value (number of pixels in between two staff lines). We use $10$ pixels for \emph{DeepScores} and $20$ pixels for \emph{MUSCIMA++}. Other than that we apply no preprocessing. We do not define a subset of target objects for our experiments, but attempt to detect all classes for which there is ground truth available. We always feed single images to the network, i.e. we only use batch size = $1$. During training we crop the full page input (and the ground truth) to patches of $960$ by $960$ pixels using random coordinates. This serves two purposes: it saves GPU memory and performs efficient data augmentation. This way the network never sees the exact same input twice, even if we train for many epochs. For all of the results described below we train individually on $loss^e$, $loss^c$ and $loss^b$ and then refine the training using $loss^{tot}$. It turns out that the prediction of $M^e$ is the most fragile, therefore we retrain on $loss^e$ again after training on the individual losses in the order defined above, before moving on to $loss^{tot}$. All the training is done using the RMSProp optimizer \cite{rmsprop} with a learning rate of $0.001$ and a decay rate of $0.995$.

\begin{table}
 \begin{center}
\begin{footnotesize}
\begin{tabular}{|rl|rl|}
\toprule
             Class &    AP@$\frac{1}{2}$ & Class &      AP@$\frac{1}{4}$ \\
\midrule
          rest16th &  0.8773 &            tuplet6 &  0.9252 \\
     noteheadBlack &  0.8619 &           keySharp &  0.9240 \\
          keySharp &  0.8185 &           rest16th &  0.9233 \\
           tuplet6 &  0.8028 &      noteheadBlack &  0.9200 \\
       restQuarter &  0.7942 &    accidentalSharp &  0.8897 \\
           rest8th &  0.7803 &           rest32nd &  0.8658 \\
      noteheadHalf &  0.7474 &       noteheadHalf &  0.8593 \\
         flag8thUp &  0.7325 &            rest8th &  0.8544 \\
       flag8thDown &  0.6634 &        restQuarter &  0.8462 \\
   accidentalSharp &  0.6626 &  accidentalNatural &  0.8417 \\
 accidentalNatural &  0.6559 &          flag8thUp &  0.8279 \\
           tuplet3 &  0.6298 &            keyFlat &  0.8134 \\
     noteheadWhole &  0.6265 &        flag8thDown &  0.7917 \\
         dynamicMF &  0.5563 &            tuplet3 &  0.7601 \\
          rest32nd &  0.5420 &      noteheadWhole &  0.7523 \\
        flag16thUp &  0.5320 &              fClef &  0.7184 \\
         restWhole &  0.5180 &          restWhole &  0.7183 \\
          timeSig8 &  0.5180 &       dynamicPiano &  0.7069 \\
    accidentalFlat &  0.4949 &     accidentalFlat &  0.6759 \\
           keyFlat &  0.4685 &         flag16thUp &  0.6621 \\
\bottomrule
\end{tabular}
\end{footnotesize}
\end{center}
 \caption{AP with overlap $0.5$ and overlap $0.25$ for the twenty best detected classes of the \emph{DeepScores} dataset.}
 \label{tab:ap_deepscores}
\end{table}

\begin{table}
 \begin{center}
\begin{footnotesize}
\begin{tabular}{|rl|rl|}
\toprule
          Class &  AP@$\frac{1}{2}$ & Class &  AP@$\frac{1}{4}$ \\
\midrule
      half-rest &  0.8981 &  whole-rest &  0.9762 \\
           flat &  0.8752 &     ledger-line &  0.9163 \\
        natural &  0.8531 &       half-rest &  0.8981 \\
     whole-rest &  0.8226 &            flat &  0.8752 \\
  notehead-full &  0.8044 &         natural &  0.8711 \\
          sharp &  0.8033 &            stem &  0.8377 \\
 notehead-empty &  0.7475 &    staccato-dot &  0.8302 \\
           stem &  0.7426 &   notehead-full &  0.8298 \\
   quarter-rest &  0.6699 &           sharp &  0.8121 \\
       8th-rest &  0.6432 &          tenuto &  0.7903 \\
         f-clef &  0.6395 &  notehead-empty &  0.7475 \\
      numeral-4 &  0.6391 &    duration-dot &  0.7285 \\
       letter-c &  0.6313 &       numeral-4 &  0.7158 \\
       letter-c &  0.6313 &        8th-flag &  0.7055 \\
       8th-flag &  0.6051 &    quarter-rest &  0.6849 \\
           slur &  0.5699 &        letter-c &  0.6643 \\
           beam &  0.5188 &        letter-c &  0.6643 \\
 time-signature &  0.4940 &        8th-rest &  0.6432 \\
   staccato-dot &  0.4793 &            beam &  0.6412 \\
       letter-o &  0.4793 &          f-clef &  0.6395 \\
\bottomrule
\end{tabular}
\end{footnotesize}
\end{center}
 \caption{AP with overlap $0.5$ and overlap $0.25$ for the twenty best detected classes from \emph{MUSCIMA++}.}
 \label{tab:ap_muscima}
\end{table}

Since our design is invariant to how many objects are present on the input (as long as their centers do not overlap) and we want to obtain bounding boxes for full pages at once, we feed whole pages to the network at inference time. The maximum input size is only bounded by the memory of the GPU. For typical pieces of sheet music this is not an issue, but pieces that use very small interline values (e.g. pieces written for conductors) result in very large inputs due to the interline normalization. At about $10.5$ million pixels even a Tesla P40 with $24$ gigabytes runs out of memory. 

\subsection{Results and Discussion}\label{subsec:results}

Tab. \ref{tab:ap_deepscores} shows the average precision (AP) for the twenty best detected classes with an overlap of the detected bounding box and ground truth of $50\%$ and $25\%$, respectively. We observe that in both cases there are common symbol classes that get detected very well, but there is also a steep fall off. The detection rate outside the top twenty continues to drop and is almost zero for most of the rare classes. We further observe that there is a significant performance gain for the lower overlap threshold, indicating that the bounding-box regression is not very accurate. 

Fig. \ref{fig:ex_detection} shows an example detection for qualitative analysis. It confirms the conclusions drawn above. The rarest symbol present, an arpeggio, is not detected at all, while the bounding boxes are sometimes inaccurate, especially for large objects (note that stems, bar-lines and beams are not part of the \emph{DeepScores} alphabet and hence don't constitute missed detections). On the other hand, staccato dots are detected very well. This is surprising since they are typically hard to detect due to their small size and the context-dependent interpretation of the symbol shape (compare the dots in dotted notes or F-clefs). We attribute this to the opportunity of detecting objects in context, enabled by training on larger parts of full raw pages of sheet music in contrast to the classical processing of tiny, pre-processed image patches or glyphs. 

The results for the experiments on \emph{MUSCIMA++} in Tab. \ref{tab:ap_muscima} and Fig. \ref{fig:ex_detection}b show a very similar outcome. This is intriguing because it suggests that the difficulty in detecting digitally rendered and handwritten scores might be smaller than anticipated. We attribute this to the fully data-driven approach enabled by deep learning instead of hand-crafted rules for handling individual symbols. It is worth noting that ledger-lines are detected with very high performance (see AP@$\frac{1}{4}$). This explains the relatively poor detection of note-heads on \emph{MUSCIMA++}, since they tend to overlap.

\begin{figure}
\begin{center}
\includegraphics[scale=0.2]{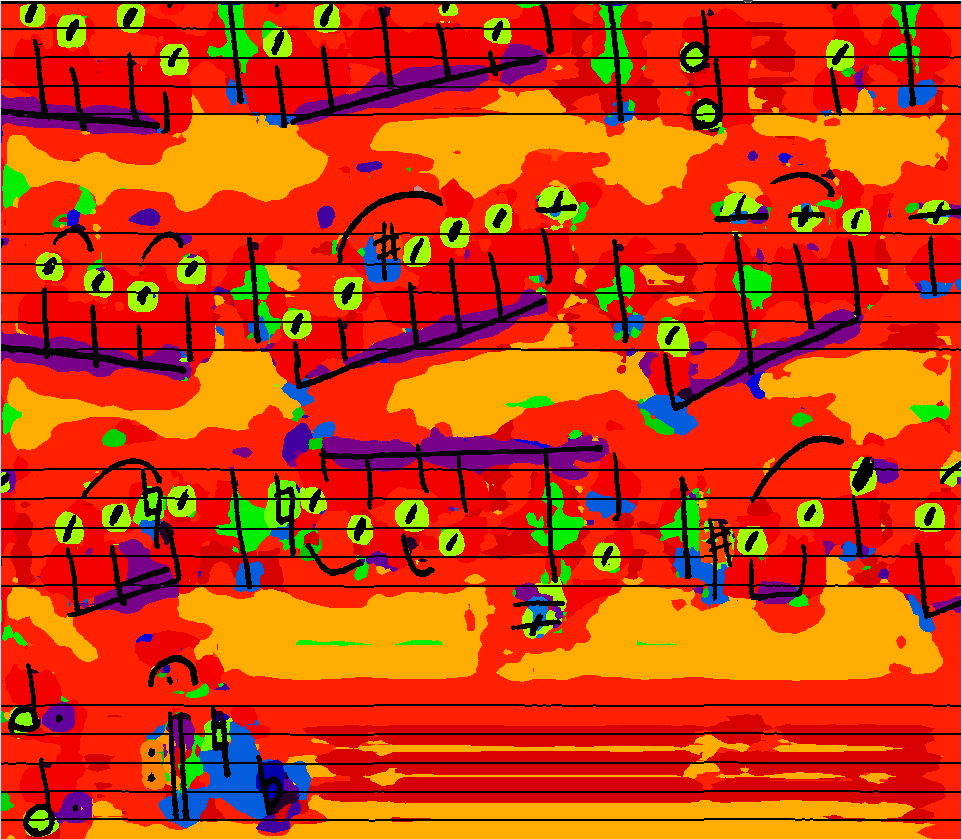}
\caption{Estimate of a class map $\hat M^c$ for every input pixel with the corresponding \emph{MUSCIMA++} input overlayed. }
\label{fig:ex_overlayed}
\end{center}
\end{figure}

Fig. \ref{fig:ex_overlayed} shows an estimate for a class map with its corresponding input overlayed. Each color corresponds to one class. This figure proofs that the network is learning a sensible representation of music notation: even though it is only trained to mark the centers of each object with the correct colors, it learns a primitive segmentation mask. This is best illustrated by the (purple) segmentation of the beams.

\section{Conclusions and Future Work}\label{sec:conc}
\begin{figure}
\begin{center}
\includegraphics[scale=0.2]{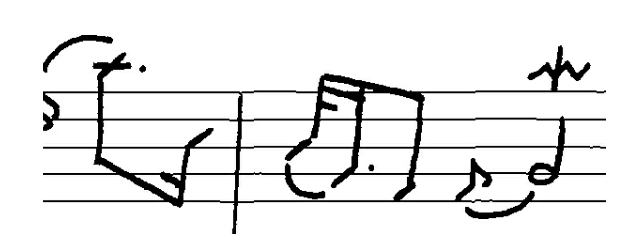}
\caption{Typical input snippet used by Pacha et al. \cite{DBLP:conf/iwdas/Pacha2018}}
\label{fig:snippet_apacha}
\end{center}
\end{figure}

We have presented a novel method for object detection that is specifically tailored to detect many tiny objects on large inputs. We have shown that it is able to detect common symbols of music notation with high precision, both in digitally rendered music as well as in handwritten music, without a drop in performance when moving to the "more complicated" handwritten input. This suggests that deep learning based approaches are able to deal with handwritten sheets just as well as with digitally rendered ones, additionally to their benefit of recognizing objects in their context and with minimal preprocessing as compared to classical OMR pipelines. Pacha et al.\cite{DBLP:conf/iwdas/Pacha2018} show that higher detection rates, especially for uncommon symbols, are possible when using R-CNN on small snippets (cp. Fig. \ref{fig:snippet_apacha}). Despite their higher scores, it is unclear how recognition performance is affected when results of overlapping and potentially disagreeing snippets are aggregated to full page results. A big advantage of our end-to-end system is the complete avoidance of error propagation in longer recognition pipeline of independent components like classifiers, aggregators etc \cite{lecun1998gradient}. Moreover, our full-page end-to-end approach has the advantages of speed (compared to a sliding window patch classifier), change of domain (we use the same architecture for both the digital and handwritten datasets) and is easily integrated into complete OMR frameworks.

\begin{figure}
\includegraphics[scale=0.2]{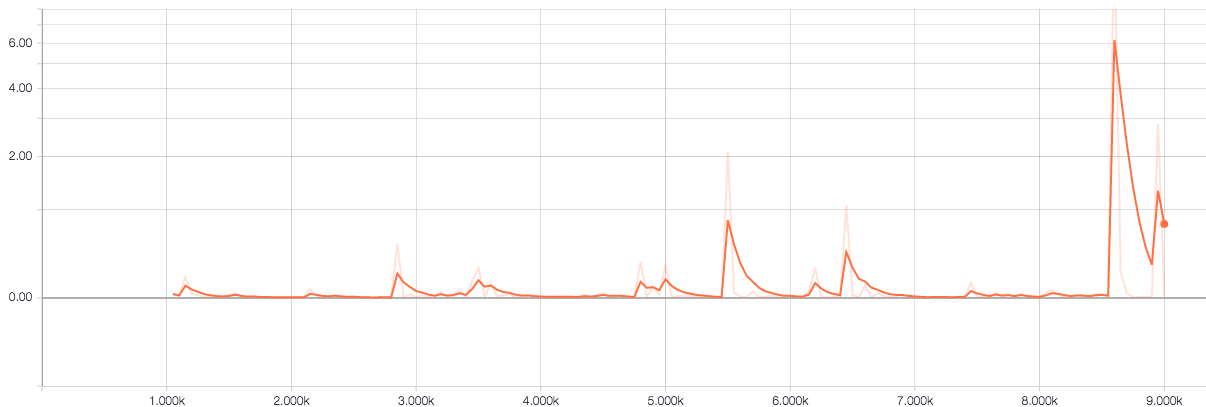}
\caption{Evolution of $loss^b$ (on the ordinate) of a sufficiently trained network, when training for another 8000 iterations (on the abscissa).}
\label{fig:loss_evolution}
\end{figure}

Arguably the biggest problem we faced is that symbol classes in the dataset are heavily unbalanced. In the \emph{DeepScores} dataset in particular, the class \emph{notehead} contains more than half of all the symbols in the entire dataset, while the top $10$ classes contain more than $85\%$ of the symbols. Considering that we did not do any class-balancing whatsoever, this imbalance had its effect in training. We observe that in cases where the symbol is common, we get a very high average precision, but it quickly drops when symbols become less common. Furthermore, it is interesting to observe that the neural network actually forgets about the existence of these rarer symbols: Fig. \ref{fig:loss_evolution} depicts the evolution of $loss^b$ of a network that is already trained and gets further trained for another $8,000$ iterations. When faced with an image containing rare symbols, the initial loss is larger than the loss on more common images. But to our surprise, later during the training process, the loss actually increases when the net encounters rare symbols again, giving the impression that the network is actually treating these symbols as outliers and ignoring them.

Future work will thus concentrate on dealing with the catastrophic imbalance in the data to successfully train DWD to detect all classes. We believe that the solution lies in a combination of data augmentation and improved training regimes (i.e. sample pages containing rare objects more often, synthesizing mock pages filled with rare objects etc.). 

Additionally, we plan to investigate the ability of our method beyond OMR on natural images. Initially we will approach canonical datasets like \emph{PASCAL VOC} \cite{DBLP:journals/ijcv/EveringhamGWWZ10} and \emph{MS-COCO} \cite{DBLP:conf/eccv/LinMBHPRDZ14} that have been at the front-line of object recognition tasks. However, images in those datasets are not exactly natural, and for the most part they are simplistic (small images, containing a few large objects). Recently, researchers have been investigating the ability of state-of-the-art recognition systems on more challenging natural datasets, like DOTA \cite{DBLP:conf/cvpr/dota}, and unsurprisingly, the results leave much to be desired. The DOTA dataset shares a lot of similarities with musical datasets, with images being high resolution and containing hundreds of small objects, making it a suitable benchmark for our DWD method to recognize tiny objects.

\vspace{0.25cm}

\newpage

\bibliography{ISMIRtemplate3}

\end{document}